\documentclass{article}

\usepackage{microtype}
\usepackage{graphicx}
\usepackage{subfigure}
\usepackage{booktabs}
\usepackage{amsmath}
\usepackage{amssymb}
\usepackage{xcolor}
\usepackage{xurl}

\usepackage{hyperref}

\usepackage[accepted]{mlsys2025}

\makeatletter
\renewcommand{\Notice@String}{}
\makeatother

\usepackage{xspace}
\usepackage{tikz}
\usetikzlibrary{arrows.meta,positioning,calc,fit,backgrounds,shapes.geometric,decorations.pathreplacing}
\usepackage{cleveref}

\newcommand{\name}{\textsc{VoxServe}\xspace}

\mlsystitlerunning{\name: Streaming-Centric Serving System for Speech Language Models}

\begin{document}

\twocolumn[
\mlsystitle{\name: Streaming-Centric Serving System for Speech Language Models}

\begin{mlsysauthorlist}
\mlsysauthor{Keisuke Kamahori}{uw}
\mlsysauthor{Rohan Kadekodi}{uw}
\mlsysauthor{Garv Ghai}{uw,gatech}
\mlsysauthor{Wei-Tzu Lee}{uw}
\mlsysauthor{Atindra Jha}{stanford}
\mlsysauthor{Stephanie Wang}{uw}
\mlsysauthor{Arvind Krishnamurthy}{uw}
\mlsysauthor{Baris Kasikci}{uw}
\end{mlsysauthorlist}

\mlsysaffiliation{gatech}{Georgia Institute of Technology}
\mlsysaffiliation{uw}{University of Washington}
\mlsysaffiliation{stanford}{Stanford University}
\mlsyscorrespondingauthor{Keisuke Kamahori}{kamahori@cs.washington.edu}

\mlsyskeywords{Speech Language Models, Machine Learning Systems, Model Serving}

\vskip 0.3in

\begin{abstract}
Speech Language Models (SpeechLMs) are increasingly deployed for streaming speech generation, where audio playback must start quickly and never stall.
Serving them requires running two very different model components on the same system. An autoregressive LLM backbone emits audio tokens, and a codec converts chunks of those tokens into waveforms.
Existing systems run the two as separately scheduled stages, and therefore either serialize them, let them interfere, or strand capacity by placing them on separate devices.
We observe that overlapping the backbone and the codec on separate parts of the GPU raises throughput.
Although this increases codec latency, most chunks can tolerate it, as long as it neither lies on the critical path of time-to-first-audio (TTFA) nor causes a playback stall.
Based on this insight, we present \name, a SpeechLM serving system that allocates GPU resources according to the urgency of chunks.
\name executes work on three asynchronous lanes. The LM lane runs the backbone on a large partition of the GPU's streaming multiprocessors, the background codec lane runs on a small partition beside it, and the urgent lane uses the whole GPU.
The scheduler queues each request to the LM lane and routes every chunk the backbone completes by its urgency. A request's first chunk and chunks close to their playback deadline go to the urgent lane, and the rest are batched on the background lane.
On Qwen3-TTS, \name sustains 1.5$\times$ the request rate of the best existing system at a 200\,ms TTFA target without playback stalls.
The code of \name is available at \url{https://github.com/vox-serve/vox-serve}.
\end{abstract}

]

\printAffiliationsAndNotice{}

\section{Introduction}

In recent years, speech models built upon large language model (LLM) foundations have made substantial progress in tasks such as Text-to-Speech (TTS) and Speech-to-Speech (STS) \cite{arora2025landscape}.
These Speech Language Models (SpeechLMs) leverage LLM backbones and neural audio codecs \cite{mousavi2025discrete} to generate and understand speech representations.
As illustrated in Figure~\ref{fig:01-speechlm-workflow}, a typical SpeechLM pipeline couples an autoregressive LLM backbone that emits discrete audio tokens with a codec that reconstructs continuous waveforms, optionally augmented by a speech encoder for audio inputs.

Serving a SpeechLM means running two very different model components in the same system.
The LLM backbone is an autoregressive decoder that advances every active request by one step at a time, much like a text LLM.
The codec (the decoder of a neural audio codec) converts each completed chunk of audio tokens into a waveform.

SpeechLMs are increasingly deployed at scale in real-world applications, including virtual assistants, content generation, and language access services \cite{openai2025realtime,elevenlabs2025intro,google2025translation}.
The proliferation of powerful open-source models has further accelerated the development of speech applications that leverage SpeechLMs \cite{amazon2025alexa,elevenlabs2025seriesc,canopylabs2025orpheus,du2024cosyvoice,wu2025step}, increasing demand for efficient SpeechLM serving \cite{peng2024survey}.

Most of these deployments require streaming, where the user starts listening while the rest of the utterance is still being generated.
Playback must start quickly, measured as \emph{Time-To-First-Audio} (TTFA), and every later audio chunk must arrive before the audio already delivered runs out, which we call \emph{streaming viability} (\S\ref{sec:background:metrics}).
A serving system must meet both for every request while serving as many requests per GPU as possible.

\begin{figure}[t]
    \centering
    \definecolor{figOneLLMFill}{HTML}{D5E8D4}\definecolor{figOneLLMDraw}{HTML}{82B366}%
\definecolor{figOneEncFill}{HTML}{F8CECC}\definecolor{figOneEncDraw}{HTML}{B85450}%
\definecolor{figOneTokFill}{HTML}{DAE8FC}\definecolor{figOneTokDraw}{HTML}{6C8EBF}%
\definecolor{figOneTxtFill}{HTML}{FFF2CC}\definecolor{figOneTxtDraw}{HTML}{D6B656}%
\begin{tikzpicture}[
    font=\sffamily\scriptsize,
    x=1cm, y=1cm,
    >={Stealth[length=1.6mm,width=1.2mm]},
    blk/.style={draw, line width=0.5pt, rounded corners=0.6pt, font=\sffamily\scriptsize\bfseries},
    flow/.style={->, line width=0.5pt},
    wave/.style={line width=0.45pt, line cap=round},
  ]
  \newcommand{\figOneWave}[5]{%
    \foreach \i in {0,...,#3}{%
      \pgfmathsetmacro{\h}{#5*(0.18 + 0.82*abs(sin(\i*97.3)*cos(\i*41.7 + 20)))}%
      \draw[wave] ({#1 + \i*#4}, {#2 - \h}) -- ({#1 + \i*#4}, {#2 + \h});
    }}

  \draw[blk, fill=figOneLLMFill, draw=figOneLLMDraw!80!black] (0,1.62) rectangle (8,2.12);
  \node[font=\sffamily\scriptsize\bfseries] at (4,1.87) {LARGE LANGUAGE MODEL};

  \draw[blk, fill=figOneEncFill, draw=figOneEncDraw] (0.05,0.92) rectangle (2.55,1.34);
  \node[font=\sffamily\scriptsize\bfseries] at (1.3,1.13) {SPEECH ENCODER};
  \draw[flow] (1.3,1.34) -- (1.3,1.62);
  \figOneWave{0.30}{0.22}{40}{0.05}{0.17}
  \node[anchor=east, inner sep=1pt] at (1.22,0.69) {Speech Input};
  \draw[flow] (1.3,0.46) -- (1.3,0.92);

  \foreach \j in {0,...,3}{%
    \draw[line width=0.4pt, fill=figOneTxtFill, draw=figOneTxtDraw]
      ({2.95 + \j*0.42}, 1.03) rectangle ++(0.3,0.3);
  }
  \draw[flow] (3.68,0.32) -- (3.68,0.92);
  \node[anchor=north, inner sep=1pt] at (3.68,0.32) {Text Input};
  \node[anchor=west, inner sep=2pt] at (3.68,0.62) {Text Tokens};

  \foreach \c in {0,1,2,4.6,5.6,6.6}{%
    \foreach \r in {0,...,3}{%
      \draw[line width=0.4pt, fill=figOneTokFill, draw=figOneTokDraw]
        ({4.21 + \c*0.42}, {2.33 + \r*0.42}) rectangle ++(0.3,0.3);
    }}
  \node[font=\normalsize] at ({4.21 + 3.3*0.42 + 0.15}, 2.96) {$\cdots$};
  \node[anchor=east, inner sep=3pt] at (4.21,2.48) {Audio Tokens};

  \draw[blk, fill=figOneTokFill, draw=figOneTokDraw!80!black] (4.11,4.04) rectangle (7.38,4.46);
  \node[font=\sffamily\scriptsize\bfseries] at ({(4.11+7.38)/2},4.25) {CODEC};
  \draw[flow] ({(4.11+7.38)/2},4.46) -- ({(4.11+7.38)/2},4.74);
  \figOneWave{4.16}{5.06}{52}{0.06}{0.26}
  \node[anchor=east, inner sep=3pt] at (4.11,5.06) {Speech Output};
\end{tikzpicture}%
    \caption{Typical workflow of SpeechLM inference. An LLM backbone autoregressively generates discrete audio tokens from text (and optionally audio) inputs, and an audio codec reconstructs continuous waveforms from those tokens.}
    \label{fig:01-speechlm-workflow}
\end{figure}

Streaming therefore gives SpeechLM serving performance requirements that differ from those of text LLM serving.
A request's \emph{first} chunk affects its TTFA, and a chunk close to its playback deadline must be decoded in time to prevent a stall, so both need low-latency processing.
Every other chunk is needed only once the audio before it has finished playing, which leaves \emph{slack} to schedule it for throughput (\S\ref{sec:background:metrics}).

These differences, between the two components and between the chunks of a request, create an opportunity that text LLM serving does not have.
We observe that, due to the relatively small sizes of modern speech models, the backbone does not always need the whole GPU to achieve high performance, and running the codec beside it on the streaming multiprocessors (SMs) that the backbone leaves idle raises throughput compared to time-slicing the GPU between the two components (\S\ref{sec:background:opportunities}).

Although each codec call incurs a longer latency with fewer SMs when run alongside the backbone, in most cases, chunks can tolerate this latency penalty. This is because most codecs neither lie on the TTFA critical path nor cause a playback stall.
Therefore, a server can split the SMs between the backbone and the codec for this common case and give the whole GPU only to the few urgent chunks.

Existing serving systems do not exploit this opportunity.
Omni-modal serving systems such as vLLM-Omni~\cite{vllm_omni2025} and SGLang-Omni~\cite{sglang_omni2025} pair a mature LLM engine~\cite{kwon2023efficient,zheng2024sglang} with a separately scheduled codec stage and forward tokens between them in fixed pipeline order, and nothing coordinates the two stages or knows which chunks are urgent (\S\ref{sec:background:existing}).

Existing systems, therefore, either disaggregate the two stages onto separate devices or colocate them on one device as black boxes that run independently, and each choice has a cost (\S\ref{sec:background:opportunities}, \S\ref{sec:design:async}).
Colocated stages either time-share the GPU which leaves SMs idle because none of the stages saturate the GPU, or run both the stages concurrently on separate streams that interfere with each other, causing a large codec batch to slow down the LLM steps running beside it, and urgent chunks to queue behind non-urgent ones.
Disaggregation avoids interference but strands capacity, because the LLM-to-codec compute ratio varies across models and workloads.
No existing system allocates GPU resources to codec work according to its urgency.

We present \name, a serving system for SpeechLMs built on the insight that GPU resources should be allocated according to the urgency of each piece of work (\S\ref{sec:design}).
On each GPU, \name executes work on three asynchronous lanes.
The LM lane runs the backbone on a large partition of the GPU's SMs, the background codec lane runs the codec on a small partition beside it, and the urgent lane runs the codec on the whole GPU at high priority.
The scheduler queues each request to the LM lane and routes every chunk generated by the backbone based on its urgency.
A request's first chunk and chunks close to their playback deadline go to the urgent lane.
The remaining chunks go to the background lane, where they wait until enough of them form an efficient batch or a playback deadline forces them out.

As a result, the backbone and the codec execute concurrently rather than in separate time slices, the first audio chunk is never queued behind background work, and codec calls are batched efficiently.
\name also briefly delays a new request's prefill when it would make a running request miss its playback deadline.

We implement \name and evaluate it on NVIDIA H100 GPUs (\S\ref{sec:eval}).
On Qwen3-TTS, at a 200\,ms TTFA target with no stalls for at least 90\% of requests, \name sustains 48 requests/s per GPU (44 at 99\% attainment), 1.5$\times$ the 32 requests/s of the best existing system, SGLang-Omni, and 1.4$\times$ its load on short utterances.
On two other models, CosyVoice~3~\cite{du2025cosyvoice3}, whose codec is a flow-matching Transformer, and Higgs Audio v3~\cite{bosonai2026higgstts3}, whose 4B backbone dominates, \name serves 30 and 32 requests/s within the same target, well above the best existing system, SGLang-Omni.
A strong synchronous design with the same kernels and prefill admission serves 38 requests/s at 90\% attainment and 30 at 99\%, against 48 and 44 for \name.

In summary, this paper makes the following contributions:
\begin{enumerate}
    \item We characterize SpeechLM serving on GPUs, showing that the backbone does not need the whole GPU, that dividing the SMs between the backbone and the codec raises throughput over time sharing but lengthens codec latency, and that urgent chunks and chunks with slack therefore need different resources (\S\ref{sec:background}).
    \item We design and implement \name, a serving system for SpeechLMs that incorporates an urgency-aware execution model that combines asynchronous lanes, SM partitions chosen among pre-built executables, a full-GPU urgent lane, slack-based codec batching, and deadline-aware prefill admission (\S\ref{sec:design}).
    \item We show that \name achieves higher SLO capacity than existing systems, with ablations attributing the gains to each mechanism (\S\ref{sec:eval}).
\end{enumerate}

\name is open source, and its code is available at \mbox{\url{https://github.com/vox-serve/vox-serve}}.

\section{Background and Motivation}
\label{sec:background}

\subsection{SpeechLM Inference}
\label{sec:background:inference}
As shown in Figure~\ref{fig:01-speechlm-workflow}, a SpeechLM consists of an LLM backbone and an audio codec~\cite{peng2024survey,arora2025landscape,mousavi2025discrete}, plus, in models that accept audio input, a speech \emph{encoder} that runs before prefill~\cite{wu2025step,zeng2024glm}.
The \emph{backbone} generates audio autoregressively, as in text generation, and each step emits one audio \emph{token} per request.
The \emph{codec}, the decoder of a neural audio codec, converts the tokens into a waveform.
To support streaming, it decodes fixed-size \emph{chunks} of tokens as soon as each is complete.
The size of each component and how much of the GPU it needs vary widely across models (Appendix~\ref{sec:appendix:models}).

\begin{figure}[t]
    \centering
    \resizebox{\linewidth}{!}{\definecolor{figTwoLM}{RGB}{46,125,50}%
\definecolor{figTwoCD}{RGB}{30,90,170}%
\definecolor{figTwoUR}{RGB}{220,95,0}%
\colorlet{figTwoLMf}{figTwoLM!22}%
\colorlet{figTwoCDf}{figTwoCD!22}%
\colorlet{figTwoURf}{figTwoUR!45}%
\colorlet{figTwoRed}{red!75!black}%
\begin{tikzpicture}[
  x=1cm, y=1cm,
  >={Stealth[length=3pt,width=2.6pt]},
  font=\fontsize{7}{8}\selectfont,
  every node/.style={inner sep=1pt},
  lm/.style={draw=figTwoLM, fill=figTwoLMf, line width=0.45pt},
  cd/.style={draw=figTwoCD, fill=figTwoCDf, line width=0.45pt},
  ur/.style={draw=figTwoUR, fill=figTwoURf, line width=0.45pt},
  rowlbl/.style={font=\fontsize{6}{7}\selectfont, text=black!70, anchor=south west, inner sep=0.5pt},
  ann/.style={font=\fontsize{6}{7}\selectfont, text=figTwoRed, align=center},
  note/.style={font=\fontsize{6}{7}\selectfont, text=black!65, align=center},
  axis/.style={->, black!55, line width=0.4pt},
  plbl/.style={font=\fontsize{7}{8}\selectfont, anchor=north},
]
\def\bh{0.15}
\def\yT{2.15}
\def\yB{1.35}
\def\W{2.6}
\newcommand{\figTwoBlk}[4]{\draw[#1] (#2,#4-\bh) rectangle (#3,#4+\bh);}
\newcommand{\figTwoSlow}[3]{%
  \draw[lm] (#1,#3-\bh) rectangle (#2,#3+\bh);
  \begin{scope}
    \clip (#1,#3-\bh) rectangle (#2,#3+\bh);
    \foreach \k in {0,...,12}{\draw[figTwoLM!70, line width=0.3pt] (#1+\k*0.08-0.3,#3-\bh) -- ++(0.3,0.3);}
  \end{scope}
  \draw[figTwoLM, line width=0.45pt] (#1,#3-\bh) rectangle (#2,#3+\bh);}

\begin{scope}[shift={(0.15,2.85)}]
  \draw[lm] (0,-0.08) rectangle (0.3,0.08);  \node[anchor=west] at (0.33,0) {LM step};
  \draw[cd] (1.7,-0.08) rectangle (2.0,0.08); \node[anchor=west] at (2.03,0) {codec batch};
  \draw[ur] (3.95,-0.08) rectangle (4.1,0.08); \node[anchor=west] at (4.13,0) {first chunk};
  \figTwoSlow{5.95}{6.25}{0}
  \node[anchor=west] at (6.28,0) {slowed LM};
\end{scope}

\begin{scope}[shift={(0,0)}]
  \node[rowlbl] at (0,\yT+\bh+0.02) {GPU};
  \figTwoBlk{lm}{0.00}{0.42}{\yT}
  \figTwoBlk{lm}{0.44}{0.86}{\yT}
  \figTwoBlk{cd}{0.88}{1.30}{\yT}
  \figTwoBlk{lm}{1.32}{1.74}{\yT}
  \figTwoBlk{cd}{1.76}{2.18}{\yT}
  \figTwoBlk{lm}{2.20}{2.60}{\yT}
  \draw[decorate, decoration={brace, amplitude=2.5pt, mirror}, figTwoRed, line width=0.45pt]
    (0.88,\yT-\bh-0.04) -- (2.18,\yT-\bh-0.04);
  \node[ann, anchor=north] at (1.53,\yT-0.27) {serialized: LM\\waits for codec};
  \draw[axis] (0,0.74) -- (\W,0.74) node[pos=1, below left=0pt, note] {time};
  \node[plbl] at (\W/2,0.52) {(a) Time sharing};
\end{scope}

\begin{scope}[shift={(2.9,0)}]
  \node[rowlbl] at (0,\yT+\bh+0.02) {stream 1};
  \node[rowlbl] at (0,\yB+\bh+0.02) {stream 2};
  \figTwoBlk{lm}{0.00}{0.42}{\yT}
  \figTwoBlk{lm}{0.44}{0.86}{\yT}
  \figTwoSlow{0.88}{1.52}{\yT}
  \figTwoBlk{lm}{1.54}{1.96}{\yT}
  \figTwoSlow{1.98}{2.60}{\yT}
  \figTwoBlk{cd}{0.92}{1.42}{\yB}
  \figTwoBlk{ur}{1.44}{1.58}{\yB}
  \figTwoBlk{cd}{2.02}{2.52}{\yB}
  \draw[->, figTwoUR, line width=0.5pt] (1.05,\yB-0.36) -- (1.05,\yB-\bh-0.02);
  \draw[<->, figTwoRed, line width=0.45pt] (1.05,\yB-0.25) -- (1.44,\yB-0.25);
  \node[ann, anchor=north] at (1.3,\yB-0.3) {urgent chunk latency};
  \node[ann] (itf) at (2.07,\yB+\bh+0.17) {interference};
  \draw[->, figTwoRed, line width=0.4pt] ([xshift=3pt]itf.north west) -- (1.3,\yT-\bh-0.01);
  \draw[->, figTwoRed, line width=0.4pt] ([xshift=-6pt]itf.north east) -- (2.25,\yT-\bh-0.01);
  \draw[axis] (0,0.74) -- (\W,0.74) node[pos=1, below left=0pt, note] {time};
  \node[plbl] at (\W/2,0.52) {(b) Colocation};
\end{scope}

\begin{scope}[shift={(5.8,0)}]
  \node[rowlbl] at (0,\yT+\bh+0.02) {LM SMs};
  \node[rowlbl] at (0,\yB+\bh+0.02) {codec SMs};
  \foreach \i in {0,...,4}{\figTwoBlk{lm}{\i*0.475}{\i*0.475+0.455}{\yT}}
  \figTwoBlk{lm}{2.375}{2.60}{\yT}
  \figTwoBlk{cd}{0.10}{1.36}{\yB}
  \figTwoBlk{ur}{1.38}{1.66}{\yB}
  \figTwoBlk{cd}{1.68}{2.60}{\yB}
  \draw[->, figTwoUR, line width=0.5pt] (0.55,\yB-0.36) -- (0.55,\yB-\bh-0.02);
  \draw[<->, figTwoRed, line width=0.45pt] (0.55,\yB-0.25) -- (1.38,\yB-0.25);
  \node[ann, anchor=north] at (0.97,\yB-0.3) {urgent chunk latency};
  \node[note, anchor=south east] at (\W,\yT+\bh+0.02) {steady steps};
  \node[note, anchor=south east] at (\W,\yB+\bh+0.02) {slow batches};
  \draw[axis] (0,0.74) -- (\W,0.74) node[pos=1, below left=0pt, note] {time};
  \node[plbl] at (\W/2,0.52) {(c) SM partition};
\end{scope}
\end{tikzpicture}%
}
    \caption{Three ways to place the LLM backbone (LM) and the codec on one GPU: (a) time sharing, (b) colocation on concurrent streams, and (c) an SM partition. In (b) and (c), urgent chunks wait behind codec batches.}
    \label{fig:02-placement}
\end{figure}

\begin{figure*}[t]
    \centering
    \begin{tabular}{@{}ccc@{}}
    \includegraphics[width=0.32\textwidth]{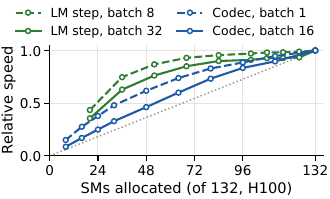} &
    \includegraphics[width=0.32\textwidth]{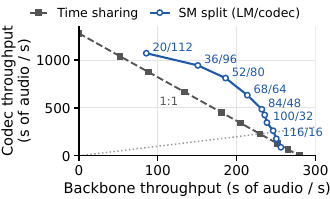} &
    \includegraphics[width=0.32\textwidth]{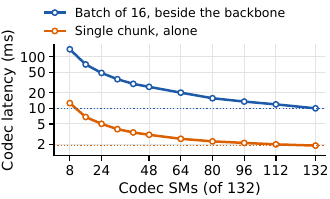} \\
    \small (a) Speed vs.\ number of SMs & \small (b) Throughput of each placement & \small (c) Codec latency on each partition \\
    \end{tabular}
    \caption{Microbenchmarks of Qwen3-TTS on one H100. (a) Speed of each component alone on an SM partition, relative to all 132 SMs. (b) Throughput of the backbone (batch 32) and the codec (batches of 16), both running back-to-back, under time sharing or on each SM partition (labels: LM/codec SMs). Below the dotted line, the codec falls behind the backbone. (c) Codec latency on each partition, and at 132 SMs alone on the whole GPU (dotted lines).}
    \label{fig:02-char}
\end{figure*}

\subsection{Metrics for Speech Serving}
\label{sec:background:metrics}

Streaming speech services are judged by how soon audio starts and whether it then plays without interruption~\cite{sierra2025voiceLatency,salesforce2025agentforce,zeghidour2025streaming,ethiraj2025toward}, rather than by the Time-To-First-Token (TTFT) and Time-Per-Output-Token (TPOT) of text LLMs~\cite{zhong2024distserve}.

\paragraph{Time-To-First-Audio (TTFA).}
The latency perceived by the client is
\begin{equation}
\mathrm{TTFA} \triangleq t_1 - t_0,
\label{eq:ttfa}
\end{equation}
where $t_0$ is the time at which the client submits the request and $t_1$ the time at which the first playable chunk is delivered.
Unlike TTFT, TTFA includes every backbone step needed to complete the first chunk and the codec call that converts it, plus the speech encoder when applicable.

\paragraph{Streaming Viability.}
Playback proceeds continuously from $t_1$, so every later chunk must arrive before the audio delivered so far has finished playing.
With $t_i$ the time at which the $i$-th chunk is delivered and $C_i$ its playback duration, uninterrupted streaming requires
\begin{equation}
t_{i+1} - t_1 \le \sum_{k = 1}^{i} C_k, \quad \forall i \ge 1.
\label{eq:sv}
\end{equation}
The constraint is cumulative, so a chunk that arrives early builds up buffered audio that absorbs a later delay.
It is also binary, since delivering a chunk before its deadline brings no perceived benefit.
A chunk's \emph{slack} is the time it can still wait before it must be decoded to meet its deadline, that is, the time until its deadline (the right-hand side of Equation~\ref{eq:sv}, measured from $t_1$) minus its decode time.
We call a chunk \emph{urgent} when it cannot wait, as is the case for a request's first chunk, which affects TTFA, and for a later chunk whose deadline is close.

\paragraph{SLO attainment and capacity.}
A request meets the per-request service-level objective (SLO) if its TTFA is at most a target $T$ and it never stalls, that is, Equation~\ref{eq:sv} holds for all of its chunks.
We set $T = 200$\,ms, following studies of turn-taking in conversation~\cite{stivers2009universals}, telephony delay guidelines~\cite{itu2003g114}, and the latency budgets of voice agents~\cite{elevenlabs2026latency}.
\emph{Attainment} at an offered request rate is the fraction of requests that meet the SLO, counting rejected and failed requests as misses, and the \emph{SLO capacity} of a system is the highest request rate at which attainment stays at or above a threshold at that rate and every lower one.
We use thresholds of 90\% and 99\%, in line with prior LLM serving work~\cite{zhong2024distserve, patel2024splitwise, agrawal2024sarathi}.

\subsection{Opportunities for Improving SLO Capacity}
\label{sec:background:opportunities}

SpeechLM serving is heterogeneous along two axes.
The components of a SpeechLM, the LLM backbone and the codec, differ in how often they run and in how much of the GPU they need.
For Qwen3-TTS, for example, the backbone runs one step for every audio token, or 80\,ms of audio, while the codec runs once per chunk of ten tokens, or 0.8\,s of audio.
Moreover, the chunks within a single request differ in how soon they must be delivered (\S\ref{sec:background:metrics}).
Together, these create opportunities for maximizing SLO capacity in SpeechLM serving.

Here, we analyze Qwen3-TTS~\cite{hu2026qwen3tts} on one H100 GPU (132 SMs) to quantify the opportunities for improving GPU utilization and SLO capacity.
First, we run each component in isolation to measure how much of the GPU it needs (O1).
Second, we run both components together under different GPU sharing strategies (O2).
Finally, we compare the codec's latency with the latency that the chunks within a request require (O3).

\paragraph{O1: High performance does not require the entire GPU.}
SpeechLM backbones and codecs have at most a few billion parameters (e.g., the backbone of Qwen3-TTS has 1.7B), while modern GPUs offer TB/s of memory bandwidth and PFLOP/s of compute.
In practice, the backbone does not need the entire device (Figure~\ref{fig:02-char}a).
On 108 of the 132 SMs, for instance, it keeps 93--99\% of its full-GPU speed.

\paragraph{O2: Overlapping the components on dedicated SMs raises throughput.}
On one GPU, the backbone and the codec can be placed in three ways (Figure~\ref{fig:02-placement}).
\emph{Time sharing} runs them one after the other.
\emph{Colocation} runs them concurrently on separate streams and lets the hardware divide the SMs between them.
An \emph{SM partition} confines each to a disjoint set of SMs, for example with NVIDIA green contexts~\cite{nvidia_green_contexts}.
With more than one GPU, the stages can also be disaggregated onto separate devices (\S\ref{sec:design:multigpu}).

Figure~\ref{fig:02-char}b plots the throughput of the backbone and the codec when both run back-to-back, under time sharing or on each SM partition.
Under time sharing, the two components' times add up, so every split of time between them gives a point on a straight line (dashed).
On separate SM partitions, the codec uses SMs that the backbone leaves idle (O1).
Almost every SM partition lies above this line, on the Pareto frontier, showing the throughput benefit of running the two components side by side on dedicated SMs.

\paragraph{O3: Urgent and slack chunks need different resources.}
Still, the throughput improvement comes at the cost of longer codec latency (Figure~\ref{fig:02-char}c).
For example, with 24 SMs, the smallest partition that keeps up with the backbone, a batch of 16 chunks takes 49\,ms beside the backbone and a single chunk 5.0\,ms, against 10 and 1.9\,ms on the whole GPU.
Chunks with slack can absorb this latency and thus run in large batches on a small partition beside the backbone.
Urgent chunks, such as the first chunk of a request, cannot (\S\ref{sec:background:metrics}); they need the whole GPU, if only for a few milliseconds, to keep TTFA low.

\paragraph{Takeaways for system design.}
To maximize throughput, a serving system should split the SMs between the backbone and the codec (O1, O2), but it must keep the resulting latency growth away from urgent chunks (O3).
Urgent chunks are few, as a Qwen3-TTS request in our LibriTTS workload (\S\ref{sec:eval:setup}) has 10.2 chunks on average, so about 90\% of chunks can have slack.
A system can therefore split the SMs for the common case and handle only the small number of urgent chunks specially, which is the design we present next (\S\ref{sec:design}).

\subsection{How Existing Systems Are Built}
\label{sec:background:existing}

SpeechLM deployments today pair a mature LLM serving engine~\cite{kwon2023efficient,zheng2024sglang,vaidya2023tensorrtllm} for the backbone with a separate pipeline for the codec and any encoder.
Omni-modal systems such as vLLM-Omni~\cite{vllm_omni2025} and SGLang-Omni~\cite{sglang_omni2025} decompose the model into stages declared in a configuration file and run each as a separately scheduled unit, often in its own process.
The backbone stage reuses the LLM engine's continuous batching, the codec runs as a lightweight streaming executor, and work moves between them in fixed pipeline order.

This design leaves the system blind to both axes of heterogeneity (\S\ref{sec:background:opportunities}).
Nothing schedules the backbone and the codec together or manages how they share the GPU, and the system cannot tell urgent chunks from chunks with slack.

\section{Design}
\label{sec:design}

\begin{figure*}[t]
    \centering
    \begin{tikzpicture}[
  x=1cm, y=1cm,
  >={Stealth[length=4pt,width=3.5pt]},
  font=\fontsize{7}{8}\selectfont,
  every node/.style={inner sep=1pt, align=center},
  region/.style={draw=black!55, rounded corners=3pt, line width=0.5pt},
  ttl/.style={font=\fontsize{8}{9.5}\selectfont\bfseries, anchor=north},
  sub/.style={font=\fontsize{6.5}{7.5}\selectfont, text=black!70},
  nextC/.style={draw=vxCD!80!black, fill=vxCD!40, line width=0.5pt},
  nextU/.style={draw=vxUR!85!black, fill=vxUR!70, line width=0.5pt},
  lmdone/.style={draw=vxLM!80!black, fill=vxLM!45, line width=0.4pt},
  lmstep/.style={draw=vxLM!70!black, fill=vxLM!45, line width=1pt},
  lane/.style={draw, rounded corners=2pt, line width=0.6pt, minimum width=1.6cm, minimum height=0.5cm},
  card/.style={draw, rounded corners=1pt, line width=0.5pt, minimum width=1.6cm, minimum height=0.5cm},
  flow/.style={->, line width=0.6pt},
  evt/.style={->, densely dashed, line width=0.5pt, black!70},
]
\definecolor{vxLM}{RGB}{46,125,50}
\definecolor{vxCD}{RGB}{30,90,170}
\definecolor{vxUR}{RGB}{220,95,0}
\colorlet{vxLMf}{vxLM!14}
\colorlet{vxCDf}{vxCD!13}
\colorlet{vxURf}{vxUR!16}
\def\vsm{\fontsize{6.5}{7.5}\selectfont}

\def\yL{5.6}
\def\yC{4.41}
\def\yU{3.22}

\draw[region, fill=black!1.5] (1.72,1.62) rectangle (11.03,6.9);
\node[ttl] at (6.38,6.9) [yshift=-2pt] {CPU};

\draw[region, fill=black!2] (11.15,1.62) rectangle (16.8,6.9);
\node[ttl] at (13.98,6.9) [yshift=-2pt] {GPU};

\draw[region, fill=black!4] (-0.2,2.25) rectangle (0.8,5.95);
\node[ttl] at (0.3,5.95) [yshift=-2pt] {Clients};
\foreach \i/\yy in {1/5.18,2/4.48,3/3.78,4/3.08}{
  \draw[black!60, fill=white, rounded corners=1pt] (-0.03,\yy-0.2) rectangle (0.63,\yy+0.2);
  \node[sub] at (0.3,\yy) {req.\,\i};
}
\foreach \d in {2.76,2.67,2.58}{\fill[black!60] (0.3,\d) circle (0.016);}
\draw[flow] (0.8,5.45) -- (1.85,5.45) node[sub, above=1pt] at (1.26,5.45) {request};
\draw[flow] (1.85,2.55) -- (0.8,2.55) node[sub, above=1pt] at (1.26,2.55) {response};

\draw[region, fill=black!3] (1.85,1.75) rectangle (8.1,6.4);
\node[ttl] at (4.98,6.4) [yshift=-2pt] {Scheduler};
\def\xo{2.66}\def\pc{0.44}
\def\lw{0.17}\def\lh{0.14}
\def\lp{0.22}
\def\cw{0.37}\def\ch{0.22}
\def\xG{6.05}
\def\xB{6.35}
\def\xw{6.65}
\def\ckx#1{\pgfmathsetmacro\cX{\xo+\pc*#1+\pc/2}}
\def\chk#1#2#3{\ckx{#1}\draw[#3] (\cX-\cw/2,#2-\ch/2) rectangle (\cX+\cw/2,#2+\ch/2);}
\def\played#1#2{\chk{#1}{#2}{black!55, fill=black!45, line width=0.4pt}}
\def\unpl#1#2{\chk{#1}{#2}{black!55, fill=black!12, line width=0.4pt}}
\def\playing#1#2#3{\unpl{#1}{#2}%
  \fill[black!45] (\cX-\cw/2,#2-\ch/2) rectangle (\cX-\cw/2+#3*\cw,#2+\ch/2);
  \draw[black!55, line width=0.4pt] (\cX-\cw/2,#2-\ch/2) rectangle (\cX+\cw/2,#2+\ch/2);
  \draw[black, line width=0.6pt] (\cX-\cw/2+#3*\cw,#2+\ch/2) -- (\cX-\cw/2+#3*\cw,#2-\ch/2-0.02);
  \fill[black] (\cX-\cw/2+#3*\cw,#2-\ch/2-0.02) -- ++(-0.055,-0.09) -- ++(0.11,0) -- cycle;}
\def\nxt#1#2#3{\chk{#1}{#2}{#3}}
\def\fut#1#2{\chk{#1}{#2}{black!50, densely dashed, line width=0.4pt}}
\def\lmx#1{\pgfmathsetmacro\lX{\xo+(\pc-\cw)/2+\lw/2+(#1)*\lp}}
\def\lmb#1#2#3{\draw[#3] (#1-\lw/2,#2-\lh/2) rectangle (#1+\lw/2,#2+\lh/2);}
\def\lmrow#1#2{\foreach \s in {1,...,#2}{\lmx{\s-1}\lmb{\lX}{#1}{lmdone}}
  \lmx{#2}\lmb{\lX}{#1}{lmstep}}
\node[sub, anchor=west] at (2.0,5.48) {req.\,1};
\lmrow{5.6}{11}
\played{0}{5.36}\playing{1}{5.36}{0.3}\unpl{2}{5.36}\nxt{3}{5.36}{nextC}\fut{4}{5.36}\fut{5}{5.36}
\node[sub, anchor=west] at (2.0,4.78) {req.\,2};
\lmrow{4.9}{7}
\playing{0}{4.66}{0.6}\unpl{1}{4.66}\nxt{2}{4.66}{nextC}\fut{3}{4.66}
\node[sub, anchor=west] at (2.0,4.08) {req.\,3};
\lmrow{4.2}{8}
\played{0}{3.96}\played{1}{3.96}\playing{2}{3.96}{0.85}\nxt{3}{3.96}{nextU}\fut{4}{3.96}
\node[sub, anchor=west] at (2.0,3.38) {req.\,4};
\foreach \d in {3.1,3.01,2.92}{\fill[black!60] (2.24,\d) circle (0.016);}
\lmrow{3.5}{2}
\nxt{0}{3.26}{nextU}\fut{1}{3.26}
\def\sw#1{\tikz[baseline=-0.6ex, sharp corners]\draw[#1] (0,-0.09) rectangle (0.24,0.1);}
\def\sl#1{\tikz[baseline=-0.6ex, sharp corners]\draw[#1] (0,-\lh/2) rectangle (\lw,\lh/2);}
\node[sub, anchor=south west, align=left, draw=black!45, fill=white, rounded corners=1.5pt, line width=0.4pt, inner sep=2pt] at (1.95,1.83)
  {LM step:\enspace \sl{lmdone}~done\enspace \sl{lmstep}~next\qquad
   \tikz[baseline=-0.6ex, sharp corners]\fill[black] (0,0.05) -- (-0.055,-0.04) -- (0.055,-0.04) -- cycle;~now playing\\[0.5pt]
   codec step:\enspace \sw{black!55, fill=black!45, line width=0.4pt}~played\enspace
   \sw{black!55, fill=black!12, line width=0.4pt}~sent, not yet played (slack)\enspace
   \sw{black!50, densely dashed, line width=0.4pt}~future\\[0.5pt]
   next codec step:\enspace \sw{nextC}~background\enspace \sw{nextU}~urgent};

\tikzset{rLM/.style={line width=0.6pt, vxLM!80!black},
         rBG/.style={line width=0.6pt, vxCD!80!black, dash pattern=on 2.4pt off 1.3pt},
         rUR/.style={line width=0.85pt, vxUR!85!black, line cap=round, dash pattern=on 0pt off 1.5pt}}
\draw[region, fill=black!3] (8.3,1.75) rectangle (10.9,6.4);
\node[ttl] at (9.6,6.4) [yshift=-2pt] {Inference engine};
\node[lane, draw=vxLM, fill=vxLMf] (lL) at (9.95,\yL) {\textbf{LM lane}};
\node[lane, draw=vxCD, fill=vxCDf] (lC) at (9.95,\yC) {\textbf{Codec lane}};
\node[lane, draw=vxUR, fill=vxURf] (lU) at (9.95,\yU) {\textbf{Urgent lane}};
\def\qr{8.95}\def\qw{0.15}\def\qp{0.2}
\def\que#1#2#3#4#5#6{%
  \draw[#2!80!black, fill=white, rounded corners=1.5pt, line width=0.5pt] (\xw,#1-0.2) rectangle (\qr,#1+0.2);
  \foreach \k in {1,...,#3}{%
    \pgfmathsetmacro\qx{\qr-0.07-\k*\qp+\qp-\qw}
    \ifnum\k>#4\relax
      \draw[#2!70, fill=#2!12, line width=0.4pt] (\qx,#1-0.1) rectangle ++(\qw,0.2);
    \else
      \draw[#2!80!black, fill=#2!50, line width=0.5pt] (\qx,#1-0.1) rectangle ++(\qw,0.2);
    \fi}
  \draw[black!75, line width=0.6pt, rounded corners=1pt]
    (\qr-0.07-#4*\qp+\qp-\qw-0.04,#1-0.15) rectangle (\qr-0.03,#1+0.15);
  \node[sub, anchor=south west, text=#2!80!black, inner sep=0.5pt] at (\xw,#1+0.2) {#5};
  \draw[flow, #6] (\qr,#1) -- (9.15,#1);}
\que{\yL}{vxLM}{8}{8}{LM queue}{rLM}
\que{\yC}{vxCD}{10}{8}{codec queue}{rBG}
\que{\yU}{vxUR}{2}{2}{urgent queue}{rUR}
\foreach \x/\w/\lab in {8.41/1.07/{KV cache}, 9.56/1.23/{codec cache}}{
  \draw[black!60, fill=white, rounded corners=1pt, line width=0.5pt] (\x,1.9) rectangle ++(\w,0.36);
  \foreach \k in {1,...,5}{\draw[black!30, line width=0.3pt] (\x+\k*\w/6,1.9) -- ++(0,0.36);}
  \node[font=\vsm, fill=white, inner sep=1pt] at (\x+\w/2,2.08) {\lab};}

\lmx{11}\draw[flow, rLM] (\lX+\lw/2+0.02,\yL) -- (\xw,\yL);
\foreach \n/\yy in {7/4.9, 8/4.2}{
  \lmx{\n}\draw[rLM] (\lX+\lw/2+0.02,\yy) -- (\xG,\yy);
  \fill[vxLM!80!black] (\xG,\yy) circle (0.035);}
\lmx{2}\draw[rLM, rounded corners=2pt] (\lX+\lw/2+0.02,3.5) -| (\xG,\yL);
\fill[vxLM!80!black] (\xG,\yL) circle (0.035);
\tikzset{halo/.style={preaction={draw=black!3, line width=2.2pt, solid, line cap=butt, -}}}
\draw[rBG, rounded corners=2pt, halo] (\xo+3.5*\pc,5.36-\ch/2) |- (\xB,5.6-0.44) -- (\xB,\yC);
\draw[black!3, line width=2.2pt, rounded corners=2pt] (\xo+2.5*\pc,4.66-\ch/2) |- (\xw,\yC);
\draw[flow, rBG, rounded corners=2pt] (\xo+2.5*\pc,4.66-\ch/2) |- (\xw,\yC);
\fill[vxCD!80!black] (\xB,\yC) circle (0.035);
\draw[rUR, rounded corners=2pt, halo] (\xo+3.5*\pc,3.96-\ch/2) |- (\xB,4.2-0.44) -- (\xB,\yU);
\draw[rUR, rounded corners=2pt] (\xo+0.5*\pc,3.26-\ch/2) |- (\xB,3.0) -- (\xB,\yU);
\draw[flow, rUR] (\xB,\yU) -- (\xw,\yU);
\fill[vxUR!85!black] (\xB,\yU) circle (0.035);

\draw[black!60, fill=white, rounded corners=2pt, line width=0.6pt] (11.27,\yU-0.3) rectangle (11.67,\yL+0.25);
\node[rotate=90] at (11.47,{(\yU-0.3+\yL+0.25)/2}) {\textbf{Hardware scheduler}};
\foreach \l/\st in {lL/rLM, lC/rBG, lU/rUR}{\draw[flow, \st] (\l.east) -- (\l.east -| 11.27,0);}

\def\gx{12.4}\def\gy{3.14}
\def\gp{0.2}\def\gb{0.15}
\def\gg{0.15}
\pgfmathsetmacro\cTop{\gy+3*\gp-\gp+\gb}
\pgfmathsetmacro\lBot{\gy+3*\gp+\gg}
\pgfmathsetmacro\lTop{\gy+11*\gp+\gg+\gb}
\pgfmathsetmacro\gR{\gx+3*\gp+\gb}
\def\gm{0.09}
\draw[rUR, rounded corners=2pt] (\gx-\gm,\gy-\gm) rectangle (\gR+\gm,\lTop+\gm);
\draw[vxLM!80!black, fill=vxLM!10, line width=0.6pt, rounded corners=1.5pt] (\gx-0.04,\lBot-0.04) rectangle (\gR+0.04,\lTop+0.04);
\draw[vxCD!80!black, fill=vxCD!10, line width=0.6pt, rounded corners=1.5pt] (\gx-0.04,\gy-0.04) rectangle (\gR+0.04,\cTop+0.04);
\foreach \c in {0,...,3}{
  \foreach \r in {0,...,2}{\draw[vxCD!70!black, fill=white, line width=0.3pt] (\gx+\c*\gp,\gy+\r*\gp) rectangle ++(\gb,\gb);}
  \foreach \r in {3,...,11}{\draw[vxLM!70!black, fill=white, line width=0.3pt] (\gx+\c*\gp,\gy+\r*\gp+\gg) rectangle ++(\gb,\gb);}}
\node[sub, anchor=south] at ({(\gx+\gR)/2},\lTop+\gm) {SMs};
\draw[flow, rLM, rounded corners=2pt] (11.67,\yL) -- (11.95,\yL) |- (\gx-0.04,{(\lBot+\lTop)/2});
\draw[flow, rBG, rounded corners=2pt] (11.67,\yC) -- (11.95,\yC) |- (\gx-0.04,{(\gy+\cTop)/2});
\draw[flow, rUR, rounded corners=2pt] (11.67,\yU) -- (11.8,\yU) |- ({(\gx+\gR)/2},\gy-\gm-0.14) -- ({(\gx+\gR)/2},\gy-\gm);

\def\tx{13.6}
\def\ts{0.03}
\foreach \s in {0,10,20,30,44,54,64,74,88}{
  \draw[vxLM!80!black, fill=vxLM!45, line width=0.3pt] (\tx+\s*\ts,\lBot) rectangle ++(10*\ts,\lTop-\lBot);
}
\foreach \s in {0,44}{
  \draw[vxCD!80!black, fill=vxCD!40, line width=0.3pt] (\tx+\s*\ts,\gy) rectangle ++(38*\ts,\cTop-\gy);
}
\fill[vxCD!40] (\tx+88*\ts,\gy) rectangle (\tx+98*\ts,\cTop);
\draw[vxCD!80!black, line width=0.3pt] (\tx+98*\ts,\gy) -- (\tx+88*\ts,\gy) -- (\tx+88*\ts,\cTop) -- (\tx+98*\ts,\cTop);
\foreach \s in {40,84}{
  \draw[vxUR!80!black, fill=vxUR!70, line width=0.3pt] (\tx+\s*\ts,\gy) rectangle ++(4*\ts,\lTop-\gy);
}
\draw[decorate, decoration={brace, amplitude=2pt}, line width=0.4pt]
  (\tx+54*\ts,\lTop+0.03) -- (\tx+64*\ts,\lTop+0.03)
  node[midway, above=1pt, font=\vsm] {one step};
\def\ya{\gy-0.2}
\draw[->, line width=0.4pt, black!70] (\tx,\ya) -- (\tx+100*\ts,\ya)
  node[sub, anchor=north east] {time};

\end{tikzpicture}
    \caption{Overview of \name. In each iteration, the per-GPU scheduler forms an LM batch and routes each request's next chunk by urgency: first chunks and chunks whose request has little slack go to the urgent lane on all SMs, and other chunks to the background codec batch.}
    \label{fig:03-system-overview}
\end{figure*}

\Cref{fig:03-system-overview} shows an overview of \name.
\name colocates the LLM backbone and the codec on a single device and gives them separate execution \emph{lanes} that run asynchronously.
On each GPU, a scheduler on a CPU thread queues work on the lanes, and the SMs are split between the backbone and the codec.
In this section, we first describe how asynchronous lanes are designed (\S\ref{sec:design:async}), then how \name achieves urgency-aware scheduling (\S\ref{sec:design:scheduling}), how it scales to distributed settings (\S\ref{sec:design:multigpu}), and how it is implemented (\S\ref{sec:design:impl}).

\subsection{Asynchronous Lanes on SM Partitions}
\label{sec:design:async}

In a naive time-sharing design, the LLM backbone and the codec take turns (Figure~\ref{fig:02-placement}a).
A naive colocation design runs them concurrently on separate streams (Figure~\ref{fig:02-placement}b) and leaves the division of the SMs to the hardware.
It therefore lacks the control that the SLO-aware scheduling of speech models needs, as an urgent chunk waits behind codec batches and a large codec batch slows the backbone steps running beside it.
Naively splitting the SMs between them (Figure~\ref{fig:02-placement}c) keeps the backbone steps steady but is also problematic for latency, as codec batches run slowly on the small partition and an urgent chunk waits behind them.
\name instead allocates SMs according to the urgency of chunks.

The scheduler drives three lanes, each of which is an in-order execution queue with an inference engine that uses different portions of GPU SMs.
The \emph{LM lane} runs backbone steps with continuous batching, as in text LLM serving systems (prefill and decode, including any depth transformer and token sampling), on a large portion of the GPU's SMs.
The \emph{background codec lane} runs batches of continuation chunks on a usually small, disjoint SM partition beside it.
The \emph{urgent lane} runs first chunks and chunks close to their playback deadline on the whole GPU at high priority.

The scheduler accepts requests from users and schedules the individual inference requests onto the three lanes.
Each user request (e.g., a text prompt for TTS) is tokenized and queued to the LM lane, and the LM's output is queued to it again until it generates an EOS token.
Once enough tokens are generated to form a chunk, the chunk is queued to one of the other lanes depending on its urgency.
If the chunk is urgent (i.e., it is the first chunk of a request or it has a small amount of slack), it is queued to the urgent lane to execute instantly.
Otherwise, the chunk is queued to the background codec lane and processed in batches to maximize throughput, since a small latency increase for such chunks does not affect end-to-end performance.

The scheduler handles each lane separately and gives it new work as soon as its previous batch finishes.
The LM lane therefore keeps issuing backbone steps while a background batch that spans several of them is still running.
When multiple lanes have pending batches, the urgent lane is given the highest priority in the GPU streams.
As a result, a long-running background lane batch does not starve an LM lane batch, and an LM lane batch does not delay an urgent chunk, ensuring low-latency inference for urgent chunks.

\name chooses the split offline.
It profiles different splits at the expected backbone and codec batch sizes at the target load, as in Figure~\ref{fig:02-char}b, and serves with the split that gives the backbone the most throughput while the codec still keeps up.
For codec-bound models, where no small partition keeps up with the backbone, this split gives the codec most of the SMs.

The chosen split and the whole GPU are prepared at server startup, and \name switches between them at runtime.
Every per-step stage is built into an executable for each batch-size bucket and SM partition, so choosing SMs is just a matter of choosing which executable to launch.
Decode steps and background batches run on the two parts of the split, while prefill and urgent batches are run on the whole GPU.

\subsection{Urgency-Aware Scheduling}
\label{sec:design:scheduling}

\paragraph{Urgency classes.}
A chunk is \emph{ready} for codec inference once all of its tokens have been generated by the LM, and has a playback deadline $d_c$, the time at which the audio delivered before it runs out (Equation~\ref{eq:sv}).
A ready chunk is \emph{urgent} if it is its request's first chunk or if $d_c - t_{\text{now}} < \delta_u$, a small margin (100\,ms by default).
All other ready chunks are \emph{background} chunks.

\paragraph{Urgent lane.}
Urgent chunks are launched as soon as they are ready.
Whenever the urgent lane has no batch in flight, the scheduler puts all urgent chunks into one batch sorted by their deadlines.
Since urgent chunks arrive one or a few at a time, the urgent lane has executables only for small batches (up to eight chunks in our implementation).
It does not wait for a running background batch but runs on all SMs beside it at a higher stream priority, with its own input and output buffers and codec scratch state.

\paragraph{Background batching within slack.}
The background lane trades the latency of chunks with slack for batch efficiency (O3).
Leaving the codec partition idle while background chunks accumulate into a batch costs little, since the backbone keeps nearly its full-GPU speed on its own partition (O1).
Once at least $B$ background chunks are ready, the scheduler launches them as one batch on the background lane, asynchronously to the LM, in deadline order and up to the largest prepared batch size.
It launches a smaller batch early when the earliest deadline is approaching, or when the LM lane has nothing left to generate.
A chunk still waiting when its deadline comes within $\delta_u$ becomes urgent and moves to the urgent lane.

Batching also works for stateful codecs, which carry per-request state from one chunk to the next.
\name keeps each request's codec state (convolution buffers, attention caches, vocoder overlap buffers) in a device-side slot, and a codec executable gathers the slots of its batch, decodes, and scatters the updated state back, so any requests can share a batch on either lane.
The scheduler keeps each request's chunks in order by dispatching its next chunk only after the previous one has finished.

\paragraph{Deadline-aware prefill admission.}
\label{sec:design:admission}
Scheduling on the LM side is another lever on TTFA and SLO attainment.
A prefill runs on the whole GPU and delays the next decode step of every running request, so admitting a new request early shortens its TTFA but can make the running ones miss their playback deadlines.
\name takes a deadline-aware approach.
It estimates the slack of each running request's next chunk, the time until its deadline minus the estimated time to generate and decode it, and defers a prefill, for at most 100\,ms in a row, while the smallest slack is below the prefill's estimated duration.
As our ablation shows (\S\ref{sec:eval:ablation}), the admission policy trades load served within the SLO against graceful degradation beyond capacity.

\subsection{Scalability}
\label{sec:design:multigpu}

The SM-partitioning idea is orthogonal to the number of GPUs.
Speech models are small enough (O1) that every GPU can fit a full copy of the backbone and the codec, so \name does not shard weights across GPUs.
It scales out by adding GPUs behind a shared front end, each with its own scheduler and engine.
Each GPU takes one of three \emph{roles}: (1) it runs both the LM backbone and the codec using the design described above; (2) it runs only the LM backbone; or (3) it runs only the codec.
In a disaggregated setup, a codec, on a partition or a whole GPU, can serve the backbones of several GPUs.

GPUs do not slow each other down, so the throughput of a \emph{placement} (an assignment of roles to GPUs) can be computed from the throughput of each role measured on its own.
A group of backbones that share one codec sustains the minimum of (1) the codec's throughput and (2) the sum of the individual backbones' throughputs.
This allows \name to rank every placement and split without running them, based on which \name decides the SM partitioning.

\begin{figure*}[t]
    \centering
    \includegraphics[width=\textwidth]{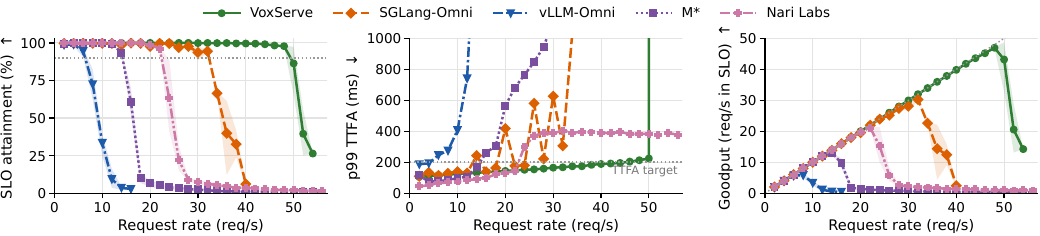}
    \caption{Serving performance for Qwen3-TTS on one H100 with LibriTTS workload. (Left) SLO attainment, the share of requests whose TTFA is at most 200\,ms and whose playback never stalls, with a dotted line at 90\%. (Middle) p99 TTFA, with the axis cut at 1\,s and a dotted line at the 200\,ms target. (Right) Goodput, the requests per second that meet the SLO, with the offered rate dotted. Lines pool three arrival seeds (median for p99), and bands show their range.}
    \label{fig:04-eval-slo}
\end{figure*}

\subsection{Implementation}
\label{sec:design:impl}

\name consists of a native Rust runtime, which serves requests, schedules, and does all per-step host work, and per-model Python programs, which load each model with PyTorch and build its stages into executables at startup.
Because the host side is a native Rust runtime, host-side scheduling overhead is minimal and takes less than 0.1\,ms at the median between inference engine execution steps.
SM partitions are implemented with green contexts~\cite{nvidia_green_contexts}, and executables are CUDA graphs captured on each partition's stream.
The engine uses FlashInfer~\cite{ye2025flashinfer} attention, \texttt{torch.compile}, and manually implemented fused kernels for the backbone layers and the codec.
Models with audio input, such as speech-to-speech models or models that clone a voice from a reference clip, run their speech encoder once per request before its prefill, on the whole GPU like prefill.
The prefill executable then reads the encoded features from a device-side pool in place of token embeddings.
A new model needs only its Python program, which declares its stages, buffers and per-request codec state, and its host-side token logic.

\section{Evaluation}
\label{sec:eval}
We answer the following questions in our evaluation:
\begin{enumerate}
\item How much load can \name serve within a streaming SLO compared with existing systems (\S\ref{sec:eval:slo})
\item How does \name{'s} performance generalize to other models (\S\ref{sec:eval:models}), larger scenarios (\S\ref{sec:eval:multigpu}), different workload patterns (\S\ref{sec:eval:workload}), and different GPU generations (\S\ref{sec:eval:gpus})
\item How do the different mechanisms contribute to overall performance in \name (\S\ref{sec:eval:ablation})
\end{enumerate}

\subsection{Setup}
\label{sec:eval:setup}

\paragraph{Hardware and models.}
Unless noted, every run uses H100 80\,GB (SXM) GPUs with 132 SMs, one GPU per server.
The main experiments use Qwen3-TTS (1.7B)~\cite{hu2026qwen3tts}, the model with the most mature streaming support in the baseline systems.
Its backbone generates 12.5 audio tokens per second of audio, each of 16 codes, the first from the 1.7B LM and the other 15 from a small depth transformer, which we count as part of the backbone rather than the codec, and a Transformer and convolutional codec turns every 10 tokens into audio.

To show generality, we also evaluate on two separate models with contrasting GPU utilization across the backbone and the codec: (\S\ref{sec:eval:models}).
CosyVoice~3~\cite{du2025cosyvoice3} pairs a smaller backbone (0.5B) with a heavier codec, a flow-matching Transformer followed by a HiFi-GAN vocoder, which dominates its GPU time.
Higgs Audio v3~\cite{bosonai2026higgstts3} pairs a larger backbone (4B), which dominates its GPU time, with a convolutional codec based on the DAC decoder.

Furthermore, we run the same evaluation on B200 GPUs (\S\ref{sec:eval:gpus}), which have 148 SMs, to show how the performance generalizes to another hardware generation with a different number of SMs.
Appendix~\ref{sec:appendix:setup} details the setup.

\paragraph{Systems.}
We compare against vLLM-Omni~\cite{vllm_omni2025} (version 0.30.0), SGLang-Omni~\cite{sglang_omni2025} (version 0.1.8), M*~\cite{jha2026mstar} (commit b37552c) and the Nari Labs Qwen3-TTS engine~\cite{narilabs2026qwen3tts} (commit e8c5b2b).
We configure vLLM-Omni and SGLang-Omni to run codec models with a chunk size of 10 tokens to align with \name.
We also increase the number of KV-cache pages of Nari Labs' engine from 256 to 1024 and run its \texttt{balanced} profile, which we found to achieve the best performance.

For Qwen3-TTS, \name runs with the 100/32 SM split, the urgent lane, background batches of at least $B = 8$ chunks and deadline-aware prefill admission (\S\ref{sec:design:scheduling}).
For CosyVoice~3 and Higgs Audio v3, it uses the 44/88 and 100/32 splits, which give CosyVoice~3's heavier codec most of the SMs, and the same $B$.
For CosyVoice~3 and Higgs Audio v3, we compare only against vLLM-Omni and SGLang-Omni, because other systems do not support these models.

\paragraph{Workload and metrics.}
We evaluate these systems with requests sampled from the test-clean split of LibriTTS~\cite{zen2019libritts} (4,837 utterances, 7.8\,s of audio on average) to simulate realistic workload in production TTS serving systems (Appendix~\ref{sec:appendix:setup}).
An open-loop client sends the speech generation requests in random order with Poisson arrivals, for 60\,s per rate in steps of 2\,req/s, and replays playback to detect stalls.

A request meets the SLO if its TTFA is at most 200\,ms and its playback never stalls (\S\ref{sec:background:metrics}), and we count rejected or failed requests as misses.
We report SLO attainment per rate and the SLO capacity, which is defined to be the highest rate before attainment falls below 90\% or 99\%, pooling the performance of three evaluations with different random seeds.

We also evaluate with the Seed-TTS~\cite{anastassiou2024seedtts} sentences of Nari Labs' benchmark, which are shorter than LibriTTS utterances (4.7\,s of audio on average), to show performance variation across workload patterns (\S\ref{sec:eval:workload}).

\subsection{SLO Capacity}
\label{sec:eval:slo}

Figure~\ref{fig:04-eval-slo} compares the serving performance for Qwen3-TTS on one H100 with LibriTTS workload. \name achieves the highest SLO capacity among the five systems.
\name can serve up to 48 req/s with SLO attainment threshold of 90\% (left figure), which is 1.5$\times$ higher than the best baseline (SGLang-Omni at 32 req/s).
At 99\% attainment, \name serves 44 req/s, against 18 for Nari Labs' engine and 12 for SGLang-Omni and M*.

While other systems like Nari Labs' engine achieve comparable TTFA as \name at low request rate (middle figure), \name can keep TTFA lower than 200ms even for higher request rate than others.
SGLang-Omni's p99 TTFA fluctuates between rates because every few thousand requests, a full Python garbage collection stops one of its server processes for 0.5--0.8\,s and delays the first audio of every request that arrives meanwhile.

Nari Labs' engine falls from 95--97\% attainment at 22 req/s to 15--32\% at 26 req/s because it completes at most about 25 req/s.
Beyond that rate, its p99 TTFA stays around 400\,ms, as its round-robin scheduler gives the prefill of new requests a turn as often as the decode of running ones.
First audio thus keeps arriving quickly, but most later chunks miss their playback deadlines, so the streams stall.

Goodput (right figure) peaks at 47 req/s for \name, against 30 for SGLang-Omni, 21 for Nari Labs' engine, 13 for M* and 6 for vLLM-Omni, and drops for every system once the offered load exceeds its capacity.

\begin{figure}[t]
    \centering
    \includegraphics[width=\columnwidth]{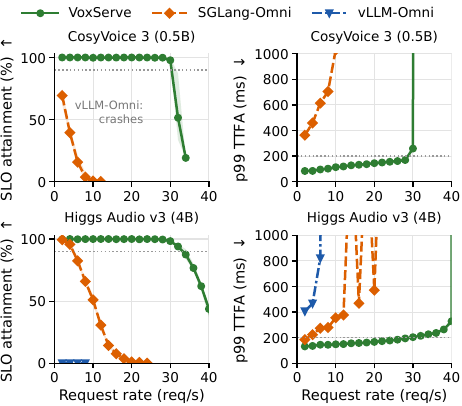}
    \caption{Serving performance for CosyVoice~3 (0.5B) and Higgs Audio v3 (4B) on one H100 with LibriTTS workload. (Left) SLO attainment. (Right) p99 TTFA, with the axis cut at 1\,s. Lines pool three arrival seeds.}
    \label{fig:04-eval-cloning}
\end{figure}

\subsection{Other Models}
\label{sec:eval:models}

Figure~\ref{fig:04-eval-cloning} compares the serving performance for CosyVoice~3 and Higgs Audio v3 against vLLM-Omni and SGLang-Omni, which support these models.
\name serves CosyVoice~3 at 30 req/s at 90\% attainment and 28 at 99\%, while SGLang-Omni meets the SLO at no tested rate (\Cref{fig:04-eval-cloning}).
At 2 req/s, its p99 TTFA is already 363\,ms and only 69\% of its requests meet the SLO.

For Higgs Audio v3, \name serves 32 req/s at 90\% attainment and 28 at 99\%, against 4 and 2 for SGLang-Omni.
vLLM-Omni meets the SLO at no tested rate for either model.
On CosyVoice~3, its codec stage crashes after 33--50 requests at 2 req/s due to a bug in version 0.30.0.

\begin{figure}[t]
    \centering
    \includegraphics[width=\columnwidth]{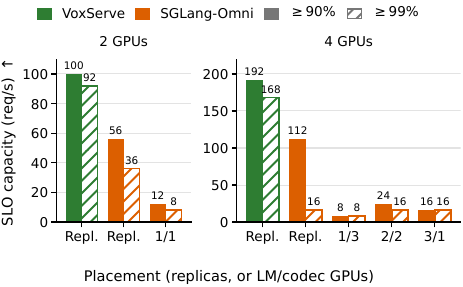}
    \caption{SLO capacity of Qwen3-TTS on two and four H100s with LibriTTS workload. SGLang-Omni runs with replication or disaggregation of LM and codec stages onto separate GPUs (LM/codec GPUs). Solid bars give the highest rate at which 90\% of requests meet the SLO, and hatched bars 99\%.}
    \label{fig:04-eval-scaling}
\end{figure}

\subsection{Scalability}
\label{sec:eval:multigpu}

\name's design is agnostic to GPU count and is scalable to distributed scenarios with multiple GPUs.
\Cref{fig:04-eval-scaling} compares its SLO capacity on two and four GPUs with that of SGLang-Omni, which runs either replicas of the same engine in data parallel manner or its LM and codec stages on separate GPUs.
\name holds the SLO up to 100 req/s at 90\% attainment (92 at 99\%) on two GPUs and 192 (168) on four, about 2 and 4$\times$ its capacity on one GPU.

SGLang-Omni scales best with pipeline replicas, which serve 56 (36) req/s on two GPUs and 112 (16) on four, so \name serves 1.8 and 1.7$\times$ as much at 90\% attainment.
SGLang-Omni's capacity at 99\% attainment stays low because of its garbage-collection pauses (\S\ref{sec:eval:slo}).

With its LM and codec stages on separate GPUs, SGLang-Omni serves at most 12 req/s on two GPUs and 24 on four, because it relays each request's codes to the codec GPU one decode step at a time, which limits one LM process to about 20 req/s.

\begin{figure}[t]
    \centering
    \includegraphics[width=\columnwidth]{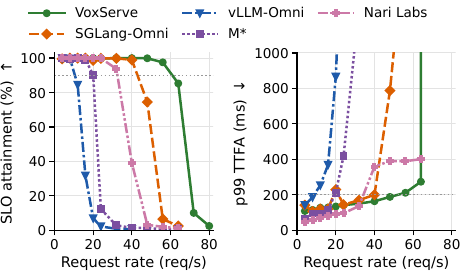}
    \caption{Serving performance for Qwen3-TTS on one H100 with Seed-TTS workload, one arrival seed per system.}
    \label{fig:04-eval-seedtts}
\end{figure}

\subsection{Another Workload}
\label{sec:eval:workload}

On the Seed-TTS sentences (\Cref{fig:04-eval-seedtts}), \name serves 56 req/s at 90\% attainment (48 at 99\%), against 40 for SGLang-Omni, 32 for Nari Labs' engine, 20 for M*, and 8 for vLLM-Omni.
These trends match the LibriTTS results (\S\ref{sec:eval:slo}), with the systems in the same order and \name serving 1.4$\times$ as much as the best baseline at 90\% attainment, which shows that its approach generalizes to another workload pattern.

\begin{figure}[t]
    \centering
    \includegraphics[width=\columnwidth]{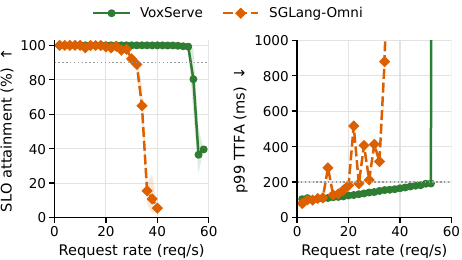}
    \caption{Serving performance for Qwen3-TTS on one B200 with LibriTTS workload. (Left) SLO attainment. (Right) p99 TTFA, with the axis cut at 1\,s. Lines pool three arrival seeds.}
    \label{fig:04-eval-b200}
\end{figure}

\subsection{Another GPU}
\label{sec:eval:gpus}

The best SM split depends on the GPU, because the backbone and the codec scale differently with SMs and memory bandwidth.
We therefore also run Qwen3-TTS on B200, which has 148 SMs and 180\,GB of HBM3e at up to 8\,TB/s, 2.4$\times$ the H100's bandwidth.
\name chooses the split in the same way as H100 (\S\ref{sec:design:async}).

On B200, it runs an 84/64 split and serves 52 req/s at both 90\% and 99\% attainment, against 30 and 10 for SGLang-Omni (\Cref{fig:04-eval-b200}).
As on the H100, \name outperforms SGLang-Omni by 1.7$\times$ at 90\% attainment, showing that its approach generalizes across GPU architectures.

\begin{figure}[t]
    \centering
    \includegraphics[width=\columnwidth]{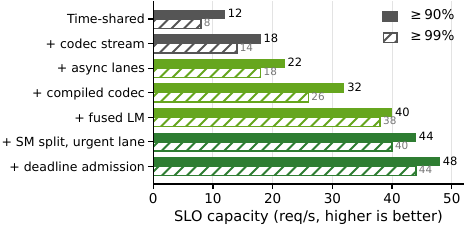}
    \caption{Ablation on Qwen3-TTS, LibriTTS. SLO capacity (TTFA $\leq$ 200\,ms, no stall) at 90\% (solid) and 99\% (hatched) attainment as \name's mechanisms are added one at a time to a synchronous loop that time-shares the GPU (three arrival seeds pooled per rate, rates in steps of 2\,req/s).}
    \label{fig:04-eval-ablation}
\end{figure}

\begin{figure}[t]
    \centering
    \includegraphics[width=\columnwidth]{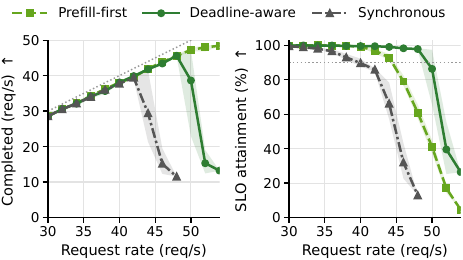}
    \caption{Qwen3-TTS on one H100 with LibriTTS workload around and past capacity, with prefill-first admission, deadline-aware admission, and the synchronous design. (Left) Completed requests per second of wall time, including the time to finish requests still running when arrivals stop, with the offered rate dotted. (Right) SLO attainment.}
    \label{fig:04-eval-overload}
\end{figure}

\subsection{Ablation Study}
\label{sec:eval:ablation}

\paragraph{Component ablation}
\Cref{fig:04-eval-ablation} adds \name's mechanisms one at a time, starting from a synchronous loop that time-shares the GPU between the backbone and the codec, which holds the SLO capacity of 12 req/s.
A second codec stream raises capacity to 18 req/s, asynchronous lanes with slack-based codec batching to 22, the compiled codec to 32, and fused backbone kernels to 40.
Splitting the SMs between the backbone and the codec, with the urgent lane for first chunks, raises capacity to 44 req/s (40 at 99\%), and deadline-aware prefill admission (\S\ref{sec:design:admission}) to 48 (44).

A synchronous design with the same kernels and admission, which issues backbone steps and codec calls from one host loop and runs the codec on a second stream, serves 38 req/s (30 at 99\%).

\paragraph{Prefill scheduling.}
Without deadline-aware admission, the LM lane admits each prefill as it arrives and decodes running requests in arrival order.
This prefill-first policy serves less load within the SLO but degrades gracefully past capacity (\Cref{fig:04-eval-overload}).
Its capacity is lower (44 vs.\ 48 req/s at 90\% attainment) because, when more requests are running than one decode batch holds, it leaves out the newest ones, which are still waiting for their first chunk.
The deadline-aware policy decodes the requests with the least slack first, so new requests go ahead of ones that already have audio buffered.
Past capacity, prefill-first admission keeps the backbone busy and still completes 48 req/s at a request rate of 52 req/s, so only TTFA grows.
Deadline-aware admission instead finds every running request short on slack and defers each prefill for the full 100\,ms, so it completes only 15 req/s.
Prefill-first admission therefore suits deployments that cannot shed or reroute excess load before it reaches the server.

\begin{figure}[t]
    \centering
    \includegraphics[width=\columnwidth]{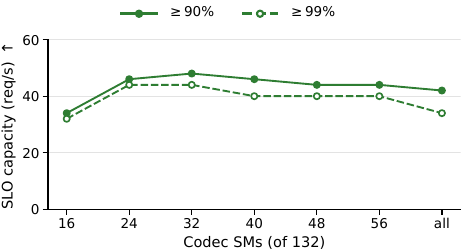}
    \caption{Sensitivity of \name to the SM split on Qwen3-TTS, LibriTTS, one H100. SLO capacity at 90\% (solid) and 99\% (dashed) attainment against the codec's SMs (``all'' shares the GPU without a split). The default is 32 SMs.}
    \label{fig:04-eval-sensitivity}
\end{figure}

\paragraph{Sensitivity to SM split.}
Near its default, \name is insensitive to the SM split (\Cref{fig:04-eval-sensitivity}).
On one GPU, the default 100/32 serves 48 req/s at 90\% attainment, and the splits from 108/24 to 76/56 serve 44--46.
116/16 leaves the codec with too few SMs and serves 34, and without a split \name serves 42.

\section{Related Work}
\paragraph{Multimodal model serving.}
LLM serving systems improve throughput and latency through KV cache management~\cite{kwon2023efficient, zheng2024sglang}, disaggregation~\cite{zhong2024distserve} and operation-level optimizations~\cite{zhu2025nanoflow}.
Multimodal serving systems place a model's stages on separate resources. EPD~\cite{singh2024efficiently} disaggregates encoding, prefill and decoding, and Cornserve~\cite{chung2026cornserve} splits any-to-any models into independently scaled components shared across applications. vLLM-Omni~\cite{vllm_omni2025} and SGLang-Omni~\cite{sglang_omni2025} extend their frameworks to omni-modal generation, and M*~\cite{jha2026mstar} executes each request as a walk through a dataflow graph of model components in one distributed runtime.
vLLM-Omni, SGLang-Omni and M* also serve SpeechLMs, but as generic multi-stage models without speech-specific optimizations.
These systems decide how stages are decomposed and placed; \name decides, within one GPU, which resources each piece of SpeechLM work receives.

Other work makes speech models cheaper to run through context compression~\cite{liu2025speech}, speculative decoding~\cite{li2025fast} or low-rank approximation~\cite{kamahori2025liteasr}, which is complementary to \name.

\paragraph{SLO-aware scheduling.}
Andes~\cite{liu2024andes} schedules text generation by quality of experience, pausing requests whose buffered text is ahead of the reader.
LiveServe~\cite{zhi2026liveserve} brings this idea to omni-modal models, prioritizing requests near a playback underrun or awaiting first audio and throttling well-buffered ones.
The Nari Labs engine~\cite{narilabs2026qwen3tts} likewise runs first audio and near-deadline chunks first.
In all of these, urgency orders work on the resources a stage already has; HorizonServe~\cite{zhang2026horizonserve} additionally caps the backbone's SMs, sized at run time from memory-bandwidth pressure.
\name instead uses each chunk's urgency to decide where it runs. Chunks with slack are batched late on a small SM partition, and urgent chunks run on the whole GPU while that partition is busy.

\section{Conclusion}
We presented \name, a serving system for streaming SpeechLMs that allocates GPU resources according to the urgency of codec work. It batches non-urgent chunks on a small SM partition beside the backbone and runs urgent chunks on the whole GPU.
Asynchronous lanes, SM partitions selected among pre-captured CUDA graphs, and per-model programs on a native runtime let it serve architecturally diverse SpeechLMs.
On Qwen3-TTS, \name sustains 1.5$\times$ the request rate of the best existing serving system at a 200\,ms TTFA target without playback stalls, and it also outperforms existing systems on two other models.

\bibliography{_ref}
\bibliographystyle{mlsys2025}

\section*{AI Usage Disclosure}
We used Claude Code (Anthropic's coding agent, with Claude Opus models) throughout this work. Under the authors' direction, it implemented parts of the system, including model ports and kernels of the runtime; launched and monitored experiments; wrote the scripts that extract the data and plot the figures; and drafted and edited parts of the text. The authors designed the system and the experiments, reviewed the code, checked results against the raw logs, and take full responsibility for the content.

\appendix
\section{Supported Models}
\label{sec:appendix:models}

Table~\ref{tab:appendix-models} lists the SpeechLMs that \name supports, and this section describes how each one maps onto the backbone, codec and encoder stages of \S\ref{sec:background:inference}.
All codecs produce 24\,kHz audio except where noted.

\begin{table*}[t]
\centering
\caption{SpeechLMs supported by \name. $K$: codes per audio token (codebooks). Chunk: tokens per codec invocation (in parentheses: audio duration).}
\label{tab:appendix-models}
\footnotesize
\setlength{\tabcolsep}{4pt}
\resizebox{\textwidth}{!}{%
\begin{tabular}{llllcrr}
\hline
Model & Backbone & Codec & Encoder & $K$ & Tokens/s & Chunk \\
\hline
Orpheus 3B~\cite{canopylabs2025orpheus}       & 3B Llama                 & SNAC (convolutional)         & --        & 7  & 82   & 28 (0.3\,s) \\
Zonos-v0.1~\cite{zyphra2025zonos}             & 1.6B                     & DAC (convolutional)          & --        & 9  & 86   & 50 (0.6\,s) \\
Higgs Audio v3~\cite{bosonai2026higgstts3} & 4B Qwen3 & Higgs codec (DAC decoder) & yes & 8 & 25 & 45 (1.8\,s) \\
CSM 1B~\cite{sesame2025uncannyvalley}         & 1B + depth transformer   & Mimi (Transformer)           & --        & 32 & 12.5 & 10 (0.8\,s) \\
Qwen3-TTS~\cite{hu2026qwen3tts}               & 1.7B Qwen3 + depth transformer & Qwen3 codec (Transformer + conv.) & --  & 16 & 12.5 & 10 (0.8\,s) \\
Voxtral-TTS~\cite{liu2026voxtraltts}                           & 4B Mistral               & Voxtral codec                & --        & 37 & 12.5 & 25 (2.0\,s) \\
CosyVoice~2~\cite{du2024cosyvoice}            & 0.5B Qwen2.5             & Flow matching + HiFi-GAN     & yes & 1  & 25   & 28 (1.1\,s) \\
CosyVoice~3~\cite{du2025cosyvoice3} & 0.5B Qwen2 & Flow matching (DiT) + HiFi-GAN & yes & 1 & 25 & 30 (1.2\,s) \\
Chatterbox~\cite{chatterbox2025}              & 0.5B Llama               & Flow matching + HiFi-GAN     & --        & 1  & 25   & 25 (1.0\,s) \\
GLM-4-Voice~\cite{zeng2024glm}                & 9B GLM-4                 & Flow matching + HiFi-GAN     & yes & 1  & 12.5 & 25 (2.0\,s) \\
Step-Audio 2 mini~\cite{wu2025step}           & 8B                       & Flow matching + HiFi-GAN     & yes & 1  & 25   & 28 (1.1\,s) \\
\hline
\end{tabular}}
\end{table*}

\paragraph{Codebook-flattening models.}
\emph{Orpheus} fine-tunes a Llama-3 3B backbone to emit the seven SNAC~\cite{siuzdak2024snac} codes of each audio token as a single stream, one code per backbone step, so it needs no depth transformer.
Its codec decodes a sliding window of four audio tokens after every token (28 codes with an overlap of 21) and keeps no state between invocations.
Each invocation is a few small convolutions, which makes Orpheus the most backbone-bound model we serve.

\paragraph{Multi-head and delay-pattern models.}
\emph{Zonos} predicts its nine DAC~\cite{kumar2023high} codebooks in parallel with nine output heads and staggers them with a delay pattern, which \name reverts before decoding.
Its prompt carries continuous conditioning features (speaker embedding, emotion, pitch and speaking rate) alongside token embeddings.
The DAC decoder runs at 44.1\,kHz on a 50-token window with a nine-token overlap, and its output is resampled to 24\,kHz.
\emph{Higgs Audio v3} also predicts its eight codebooks in parallel with a delay pattern, from one output head that shares its weights with the codebook embedding, and clones a voice from a reference clip encoded into the same codes.
Its decoder is non-causal, so \name decodes 25-token chunks with 12 tokens of left context and 8 tokens of look-ahead.

\paragraph{Depth-transformer models.}
\emph{CSM} and \emph{Qwen3-TTS} predict the first codebook of each token with the backbone and the remaining codebooks (31 and 15) with a small depth transformer that runs once per codebook inside every backbone step.
The depth transformer has its own short KV cache, which is reset every step, and its sampled codes are summed into the next backbone input.
CSM decodes ten-token chunks with Mimi~\cite{defossez2024moshi} and applies an audio watermark.
Qwen3-TTS decodes ten-token chunks with a causal Transformer-plus-convolution codec whose per-request state (a 72-token sliding-window attention cache, convolution caches for the pre-convolution, upsampling and decoder blocks, and a position offset) carries across chunks.
Qwen3-TTS also supports input streaming, in which text tokens are injected into the backbone input as they arrive.

\paragraph{Inline acoustic head.}
\emph{Voxtral-TTS} emits one semantic and 36 acoustic codes per token.
Instead of a depth transformer, a small flow-matching head runs seven Euler steps with classifier-free guidance on the backbone's hidden state to produce the acoustic codes, inside the backbone step.
Its codec decodes 25-token chunks with 25 tokens of left context carried as state, and the first chunk is shortened to five tokens to reduce TTFA.

\paragraph{Flow-matching codecs.}
\emph{CosyVoice~2}, \emph{Chatterbox}, \emph{GLM-4-Voice} and \emph{Step-Audio~2} emit a single stream of semantic speech tokens from a small or medium backbone, and convert them to audio with a conditional flow-matching model followed by a HiFi-GAN vocoder~\cite{kong2020hifi}.
The flow model runs several denoising steps over a Transformer and convolution estimator for every chunk, which makes these codecs far more expensive than the backbone (over 90\% of the GPU time when the two run one after the other).
CosyVoice~2 and Step-Audio~2 carry streaming state across chunks (encoder and estimator attention and convolution caches, and vocoder overlap buffers), whereas Chatterbox and GLM-4-Voice decode each chunk independently.
\emph{CosyVoice~3} replaces the estimator with a 22-block diffusion Transformer and clones a voice from a reference clip and its transcript.
\name decodes it in chunks of 25 tokens, each becoming one block of the model's block-causal flow, and runs its vocoder incrementally, carrying the source phase across chunks.
GLM-4-Voice and Step-Audio~2 are speech-to-speech models: a speech encoder turns input audio into discrete tokens (GLM-4-Voice) or continuous features (Step-Audio~2) before prefill, and the backbone interleaves text and audio tokens in its output.
Chatterbox applies an audio watermark.

\section{Experimental Setup}
\label{sec:appendix:setup}

This appendix describes how every run behind the numbers in the introduction, \S\ref{sec:background:opportunities} and \S\ref{sec:eval} was set up.

\paragraph{Common setup.}
All runs use one node with four NVIDIA H100 80\,GB HBM3 (SXM) GPUs, each with 132 SMs, driver 590.48.01.
GPUs are allocated exclusively; a run gets one GPU (two or four for the multi-GPU runs) and, in most runs, 8--16 CPU cores and 96--192\,GB of host memory.
\name runs its native (Rust) runtime with Python only for model loading, graph capture and request preprocessing, on PyTorch~2.8.0 with CUDA~12.8 and FlashInfer~0.2.11.
The B200 runs of \S\ref{sec:eval:gpus} use one GPU of a separate server.
Weights and activations are BF16.
Every stage is captured as a CUDA graph per batch size: decode and codec graphs at powers of two up to the maximum batch, urgent-lane graphs at batch 1, 2, 4 and 8, and prefill graphs for 256 to 2048 tokens.
Decode and codec graphs are captured once per split, prefill and urgent-lane graphs only on the whole GPU.
For Qwen3-TTS on the 100/32 split, this gives 36 graphs, captured in about three minutes at startup, most of it compiling the codec.
The KV cache uses 64-token pages and up to 2048 tokens of context per request.

\paragraph{Serving workload.}
All serving runs drive the server's streaming endpoint with the same open-loop client.
The LibriTTS runs use the normalized text of the 4,837 utterances of the LibriTTS test-clean split~\cite{zen2019libritts}, of 1--71 words and 18.7 on average.
Each arrival seed draws its own order of the utterances as well as its own arrival times, and a discarded burst of 64 concurrent requests precedes each sweep.
A sweep stops after a rate at which fewer than half of the requests receive audio within 1\,s.
Arrival times are drawn from a Poisson process with a fixed seed and kept on an absolute schedule, so a slow server does not lower the offered load.
Each rate runs for a 60\,s arrival window, after which the client waits until every request finishes or times out after 900\,s.
The rates of one sweep run back to back against the same server, after four warm-up requests that are not counted.

\paragraph{Metrics.}
The client receives 16-bit PCM at 24\,kHz and timestamps every network chunk.
TTFA is the time from sending the request to receiving the first audio bytes, so it includes HTTP and queueing.
The client models a player that starts at the first byte and pauses whenever the next audio has not arrived, and a request \emph{stalls} if these pauses add up to 10\,ms or more.
A request meets the SLO if it completes, its TTFA is at most $T$ (200, 250, 500 or 1000\,ms; the paper uses 200\,ms), and it does not stall.
Rejected requests (SGLang-Omni returns HTTP 503 once its queue of 64 is full) and failed requests count as misses.
Attainment and SLO capacity are as defined in \S\ref{sec:background:metrics}, over the tested rates only.

\paragraph{Baselines.}
vLLM-Omni~0.30.0 (vLLM~0.30.0) uses its shipped Qwen3-TTS deployment: Model Runner V2, asynchronous chunk streaming, the backbone and the codec as two stages on the same GPU with 30\% of its memory each, \texttt{max\_num\_seqs} 64, codec batches of at most 4, and the same backbone sampling parameters as \name.
SGLang-Omni~0.1.8 (SGLang~0.5.21) uses its shipped Qwen3-TTS pipeline with \texttt{max\_running\_requests} and \texttt{max\_queued\_requests} of 64 and its default chunk ramp.
Both frameworks run PyTorch~2.13 built for CUDA~13.0, with the CUDA~13.1 toolkit for their JIT kernels.
M* runs from its main branch (commit b37552c) with \texttt{mstar serve qwen3\_tts\_1p7b} and its default configuration, which caps batches at 32 and splits the text into sentences on the server.
The Nari Labs engine (commit e8c5b2b) is installed with its pinned dependencies and serves the model revision its benchmark reports use (0c0e305).
Its \texttt{balanced} profile schedules round robin with stage batches of up to 64 and codec chunks of 4, 4, 8, 16 and then 25 tokens; its \texttt{ttfa} profile schedules by deadline (1\,s lead) with batches of up to 32 and chunks of 1, 2, 4, 8 and then 12 tokens.
Both admit up to 128 active requests and reject further ones with HTTP 429.
Both reserve a KV cache of 256 pages of 128 tokens and stop the engine with an error when it is exhausted.
The balanced profile serves more load than the ttfa profile on the Seed-TTS sentences (32 vs.\ 20 req/s at 90\% attainment), so \S\ref{sec:eval} reports it.

\paragraph{CosyVoice~3 and Higgs Audio v3.}
The runs of \S\ref{sec:eval:models} use the checkpoints \texttt{FunAudioLLM/\allowbreak Fun-\allowbreak CosyVoice3-\allowbreak 0.5B-\allowbreak 2512} (revision 29e01c4) and \texttt{bosonai/\allowbreak higgs-tts-3-4b} (revision 239f63f), with the same frameworks, client and workload as the Qwen3-TTS runs.
Every request carries the same reference: LibriSpeech utterance 1272-135031-0017 (6.9\,s, 16\,kHz) as a base64 WAV in \texttt{ref\_audio} and its transcript in \texttt{ref\_text}.
Each model samples with its deployment's default parameters.

vLLM-Omni~0.30.0 runs CosyVoice~3 with its shipped \texttt{cosyvoice3.yaml}.
For Higgs Audio v3, the default deployment is the high-throughput profile (\texttt{higgs\_\allowbreak multimodal\_\allowbreak qwen3\_\allowbreak high\_\allowbreak throughput.yaml}).
SGLang-Omni~0.1.8 runs both models with its default pipeline configurations, with no configuration file.

\end{document}